%% file: main.tex
\documentclass{article}
\usepackage{spconf,amsmath,graphicx}

\usepackage{hyperref}       % hyperlinks
\usepackage{url}            % simple URL typesetting
\usepackage{booktabs}       % professional-quality tables
\usepackage{amsfonts}       % blackboard math symbols
\usepackage{nicefrac}       % compact symbols for 1/2, etc.
\usepackage{microtype}      % microtypography
\usepackage{xcolor}         % colors
\usepackage{graphicx}
\usepackage{amsmath}
\usepackage{multirow}
\usepackage[font=small,labelfont=bf]{caption}
\usepackage{subcaption}

\usepackage{newfloat}
\usepackage{floatrow}
\usepackage{float}

\newfloatcommand{capbtabbox}{table}[][\FBwidth]
\usepackage{cite}

\newcommand{\todokan}[1]{#1}

\newcommand{\cc}[1]{\textcolor{red}{\emph{[claire: #1]}}}

\title{Neural Adapters for Personalization in RNN-T\cc{Train One, Get Many: Neural Adapters for Personalized Speech Recognition}\todokan{Adapting Pretrained neural-transducers for personalized speech recognition}}
\title{Contextual Adapters for Personalized Speech Recognition \\in Neural Transducers}
\name{\begin{tabular}{c}Kanthashree Mysore Sathyendra, Thejaswi Muniyappa, Feng-Ju Chang, Jing Liu, \\ Jinru Su, Grant P. Strimel, Athanasios Mouchtaris, Siegfried Kunzmann\end{tabular}}

\address{Amazon Alexa}

\begin{document}
\topmargin=0mm
\ninept
\maketitle
\begin{abstract}
Personal rare word recognition in end-to-end Automatic Speech Recognition (E2E ASR) models is a challenge due to the lack of training data. A standard way to address this issue is with shallow fusion methods at inference time. However, due to their dependence on external language models and the deterministic approach to weight boosting, their performance is limited. In this paper, we propose training neural \textit{contextual adapters} for personalization in neural transducer based ASR models. Our approach can not only bias towards \todokan{user-defined words}, but also has the flexibility to work with pretrained ASR models. Using an in-house dataset, we demonstrate that contextual adapters can be applied to any general purpose pretrained ASR model to improve \todokan{personalization}. Our method outperforms shallow fusion, while retaining functionality of the pretrained models by not altering any of the model weights. We further show that the adapter style training is superior to full-fine-tuning of the ASR models on \todokan{datasets with user-defined content}.

\end{abstract}

%\vspace{-6pt}
\section{Introduction}
\vspace{-6pt}
\label{sec:intro}
\input{sec_introduction.tex}
\section{Neural Transducers}
\input{sec_background.tex}

\section{Contextual Adapters}

\label{sec:model}
\input{sec_approach.tex}

\vspace{-15pt}
\section{Experiments}

\label{sec:exp}
\input{sec_experiments.tex}

\vspace{-6pt}
\section{Results}
\vspace{-6pt}
\label{sec:results}
\input{sec_results.tex}

\vspace{-8pt}
\section{Conclusion}
\vspace{-8pt}
In this paper, we introduced \textit{contextual adapters} to adapt pretrained RNN-T and C-T to improve speech recognition of \todokan{contextual} entities such as contact names, device names etc. While previous works have focused on training contextual models from scratch, our approach can adapt an already trained ASR model, by training only a small set of parameters. We also show that adapter-style training improves over shallow fusion baselines, while retaining the same flexibility as shallow fusion. Our approach improves by over 31\% when compared to our baseline models on a dataset containing \todokan{contextual} utterances, while degrading less than 3.5\% relative on the \todokan{non-contextual} utterances. We further demonstrate our model can handle multiple catalog types with the same kind of improvements.

\bibliographystyle{IEEEbib}
\bibliography{strings,refs}

\end{document}

%% file: sec_introduction.tex
End-to-end (E2E) ASR systems are gaining popularity due to their monolithic nature and ease of training, making them promising candidates for deployment in commercial voice assistants (VAs). While these models outperform traditional hybrid ASR models on generic speech datasets, they still struggle to recognize difficult, uncommon terms such as contact names, proper nouns and other named entities~\cite{sainath2018no, bruguier2016learning}. To provide the best user experience and recognize requests correctly, a voice assistant should be able to adapt well to each user's custom environment and preferences, and use them to improve recognition of personalized requests. Examples of personalized requests include \textit{``call [Contact Name]"}, and \textit{``turn on [Device Name]"}. 

Prior works to address this problem broadly fall into two categories: post-training integration of external language models (LMs) and training-time integration of personalized context \cite{pundak2018deep, bruguier2016learning, bruguier2019phoebe, gourav2021personalization,jain2020contextual,le2021deep, le2021contextualized}. Mainstream approaches in the first category are shallow fusion (SF)~\cite{Zhao2019,gourav2021personalization} and on-the-fly (OTF) rescoring~\cite{he2019streaming}, which construct $n$-gram finite state transducers (FSTs) based on separately trained user-dependent LMs, and boost the scores of \todokan{user-defined contextual entities (henceforth `contextual entities')}. While this approach can be applied with many pretrained ASR models, its performance is sensitive to the weights of contextual LMs~\cite{Zhao2019}, and may overboost \todokan{these contextual entities} resulting in performance degradation. In the second category, popular approaches include neural contextual biasing as performed in LAS models~\cite{pundak2018deep, 44926} where bias phrases and/or contextual entities are encoded, and the ASR model is made to bias toward them via a location-aware attention mechanism~\cite{chorowski2015attention, chen2019joint,bruguier2019phoebe}. For RNN-T models,~\cite{Zhao2019} and~\cite{le2021deep} introduced shallow fusion and deep personalized LM fusion, respectively. Both~\cite{jain2020contextual} and~\cite{chang2021context} introduced the neural contextual biasing for open domain ASR with RNN or transformer transducers. However, these approaches train the \todokan{contextual} model from scratch, and do not explore contextual adaptation of already trained models.

In this paper, we propose training lightweight \emph{contextual adapter} networks~\cite{rebuffi2017learning, rebuffi2018efficient,houlsby2019parameter} to augment pretrained neural sequence transducers, such as the RNN-Transducer (RNN-T)~\cite{graves2012sequence} and the Conformer-Transducer (C-T)~\cite{gulati2020conformer}, to improve recognition of \todokan{contextual entities} to support \todokan{personalization for users}. Specifically, the proposed contextual adapter network consists of a catalog encoder and an attention-based biasing adapter. The catalog encoder encodes \todokan{contextual entities} such as users' contact names, device names etc. into embeddings. The biasing adapter measures the correlation between the pretrained model’s intermediate representations -- such as encoder, prediction network, or joint network outputs -- and the context embeddings, to determine the \todokan{contextual entities} to attend over.

\begin{figure}[t]
    \centering
    \includegraphics[width=0.6\linewidth]{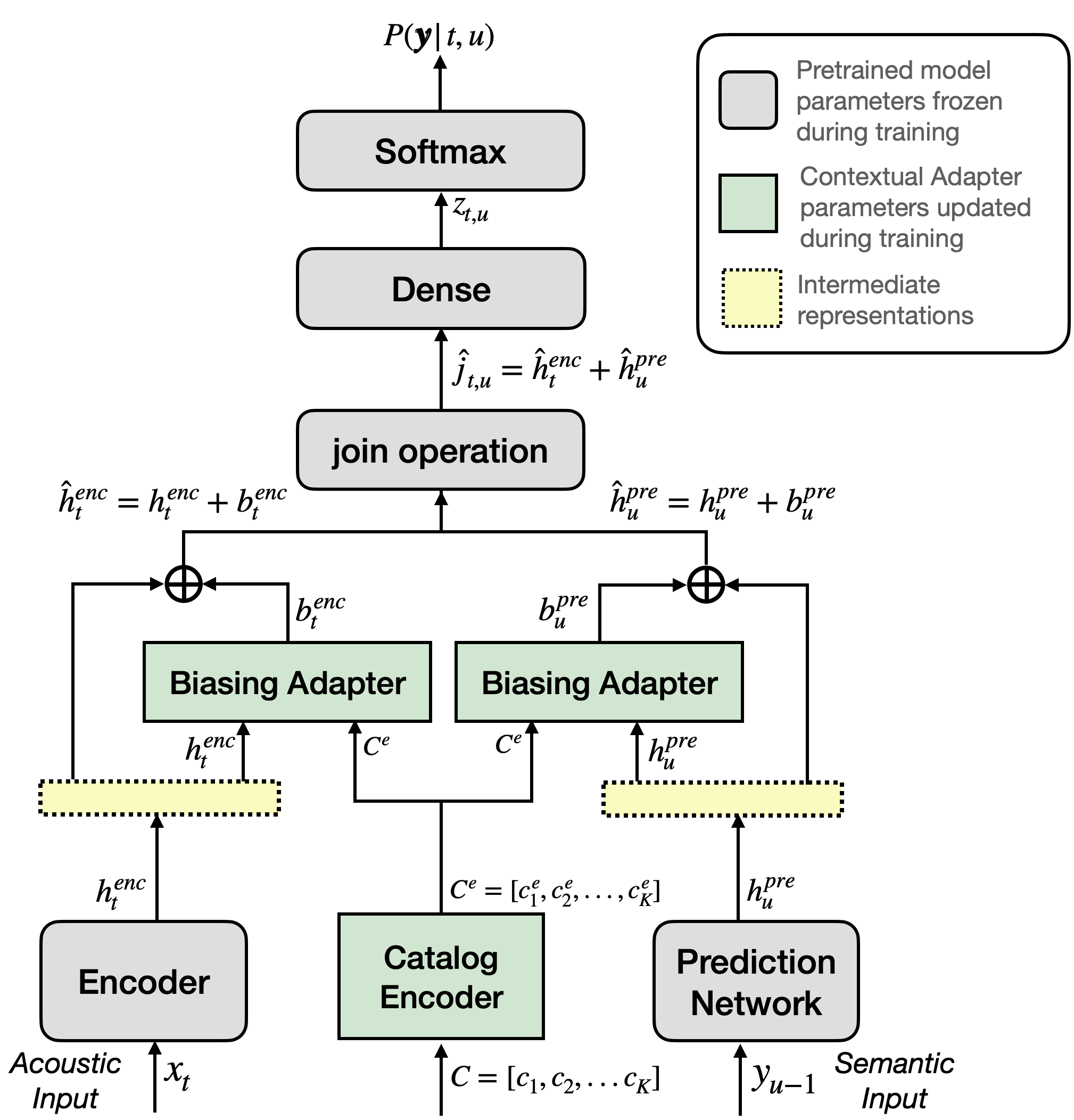}
    \vspace{0pt}
    \caption{\textbf{Contextual Adapters.} A personalized neural transducer with the proposed contextual adapters (Enc-Pred Query).}
    \label{fig:RnntAdapter}
    \vspace{-6pt}
\end{figure}

In addition to achieving personalization, the proposed contextual adapters approach has several advantages. First, it is data-efficient and requires only a small amount of \todokan{contextual datasets} for training the adapters. Second, training adapters is faster and cheaper since only a small fraction of the parameters are trained (training time reduced by \textgreater 86\% when compared to training from scratch). Third, this design has the advantage that it can easily utilize well-trained, generic ASR models and still improve personalized word prediction. Last, they offer flexibility. The same pretrained model can be adapted to recognize multiple personalized domains such as contact names, device names, or rare words, just by changing the adapter being used; thus bridging the gap between benefits of lightweight, deterministic approaches like SF and the modeling capacity of costly, fully neural methods trained from scratch. 

Using an in-house far-field dataset, we demonstrate that our method, when applied to RNN-T and C-T, outperforms the unadapted model and shallow fusion in terms of the named-entity word error rate. Furthermore, the \todokan{the data-efficiency of adapter style training makes it} attractive for low-resource settings. Finally, we show that a single contextual adapter can be generalized to recognize multiple types of \todokan{contextual} entities such as \todokan{proper} names, locations and devices.

%% file: sec_background.tex
\newcommand{\cnemb}{c^e_i}
\newcommand{\cn}{c_i}
\newcommand{\cni}[1]{c_{#1}}
\newcommand{\cniemb}[1]{c_{#1}^e}
\newcommand{\call}{C}
\newcommand{\cemball}{C^e}
\newcommand{\attni}{\alpha_i}
\newcommand{\attnit}{\alpha_i^t}
\newcommand{\dotprod}{\cdot}
\newcommand{\rootd}{\sqrt{d}}
\newcommand{\querylin}{\boldsymbol{W}^{q}}
\newcommand{\keylin}{\boldsymbol{W}^{k}}
\newcommand{\vallin}{\boldsymbol{W}^v}
\newcommand{\accousticquery}{q}
\newcommand{\semanticquery}{h_u^{pre}}
\newcommand{\encattncont}{b_t^{enc}}
\newcommand{\biasvector}{b}
\newcommand{\encbiasvector}{b^{enc}_t}
\newcommand{\decbiasvector}{b^{pre}_u}
\newcommand{\jointbiasvector}{b_{t,u}}
\newcommand{\jhat}{\hat{j}_{t,u}}
\newcommand{\maxcatsize}{K}
\newcommand{\htcap}{\hat{h}_t^{enc}}
\newcommand{\hucap}{\hat{h}_u^{pre}}
\newcommand{\nobias}{\textit{\textless no\_bias\textgreater} }
\newcommand{\encout}{h_t^{enc}}
\newcommand{\decout}{h_u^{pre}}
\newcommand{\decinp}{y_{u-1}}
\newcommand{\encinp}{x_{t}}
\newcommand{\encinpstart}{x_{0}}
\newcommand{\decinpstart}{y_{0}}
\newcommand{\catlen}{K}
\newcommand{\catembsize}{D}
\newcommand{\fullcatalog}{$\call = [\cni{1}, \cni{2}, \dots, \cni{K}]$}
\newcommand{\fullcatalogemb}{$\cemball = [\cniemb{1}, \cniemb{2} \dots \cniemb{K}]$}

Neural sequence transducers are a type of streaming E2E ASR models \cite{graves2012sequence}, typically consisting of an encoder network, a prediction network and a joint network. The encoder network produces high-level representations $\encout$ for the audio frames $\mathbf{x_{0,t}}= (\encinpstart \dots \encinp)$. The prediction network encodes the previously predicted word-pieces $\mathbf{y_{0,u-1}}=(\decinpstart \dots \decinp)$ and produces the output $\decout$. The encoder and the prediction network typically are stacked RNN layers \cite{graves2012sequence} or stacked conformer blocks~\cite{li2021better,gulati2020conformer}.
\begin{align*}
    \encout & = \text{Encoder}(\mathbf{x_{0,t}}); \quad
    \decout = \text{PredictionNetwork}(\mathbf{y_{0,u-1}}) 
\end{align*}
The joint network first fuses $\encout$ and $\decout$ via the join operation. This is passed through a series of dense layers with activations (denoted by $\phi$), then a softmax is applied to obtain the probability distribution over word-pieces as in equation~\ref{eqn:posterior} (includes the $blank$ symbol).
\vspace{-15pt}
\begin{align}
    P(y_u|t,u) = \text{softmax}(z_{t,u})    
    \label{eqn:posterior}
\end{align}
The entire model is trained with the RNN-T loss using the forward-backward algorithm that accounts for all possible alignments between the $T$ acoustic frames and $U$ word-pieces in the ground truth \cite{graves2012sequence}.

Intuitively, the encoder network is expected to behave as an acoustic model (AM) and the prediction network as a LM \cite{he2019streaming}. 

%% file: sec_approach.tex
The contextual adapters comprise two components -- a catalog encoder and biasing adapters. Given a pretrained ASR, they adapt this model to perform contextual biasing based on \todokan{contextual} catalogs. 
\vspace{-3pt}
\subsection{Catalog Encoder}
\vspace{-3pt}
The catalog encoder encodes a catalog (or list) of \todokan{contextual entities or catalogs}, \fullcatalog, producing one encoded representation or ``\todokan{entity} embedding" per entity in the catalog. With $\catlen$ entities in the user's catalog, and an embedding size of $\catembsize$, the catalog encoder outputs $C^e \in \mathbb{R}^{K\times D} $. Each entity $c_i$, is first split into fixed-length word-pieces using a sub-word tokenizer \cite{sennrich-etal-2016-neural, kudo2018subword}, then passed through an embedding lookup followed by BiLSTM layers. We use the same tokenizer as used by the RNN-T's prediction network in order to maintain compatibility between the output vocabulary and the catalog encoder. 
The final states of the BiLSTMs are forwarded as the embeddings of the named entities, denoted as \fullcatalogemb ~where
\begin{align*}
    \cnemb = \text{BiLSTM}(\text{Embedding}(c_i))
\end{align*}
Since the \todokan{catalog entities} may not always be relevant (for example, utterances in the \textit{Weather} domain may not need adapters biasing towards contact names), we also introduce a special \nobias token into our catalog (as in \cite{pundak2018deep}). 

\vspace{-3pt}
\subsection{Biasing Adapters}
\vspace{-3pt}
The biasing adapter adapts intermediate representations from the neural transducers by incorporating biasing information as shown in Figure~\ref{fig:RnntAdapter} and ~\ref{fig:ad_vars}. We propose cross-attention based biasing adapters (Figure~\ref{fig:ad_vars}~(b)) to attend over $\cemball$ based on the input query $q$. The query and the catalog entity embeddings are first projected via $\keylin$ and $\querylin$ to dimension $d$. And the attention score $\attni$ for each catalog entity embedding $\cnemb$ is computed by the scaled dot product attention mechanism~\cite{NIPS2017_3f5ee243}: 
\begin{align*}
    \attni = \text{Softmax}_i \left(\frac{\keylin \cemball  \dotprod \querylin\accousticquery}{\rootd} \right)
\end{align*}
The attention scores are used to compute a weighted sum of the value embeddings obtained by a linear projection of the catalog entity embeddings, which becomes the biasing vector, $\biasvector = \sum_{i}^{\maxcatsize} \attni ~ \vallin\cnemb$.

% Figure
% \begin{minipage}
  \begin{figure}[tb]
    \centering
    \includegraphics[width=0.93\linewidth]{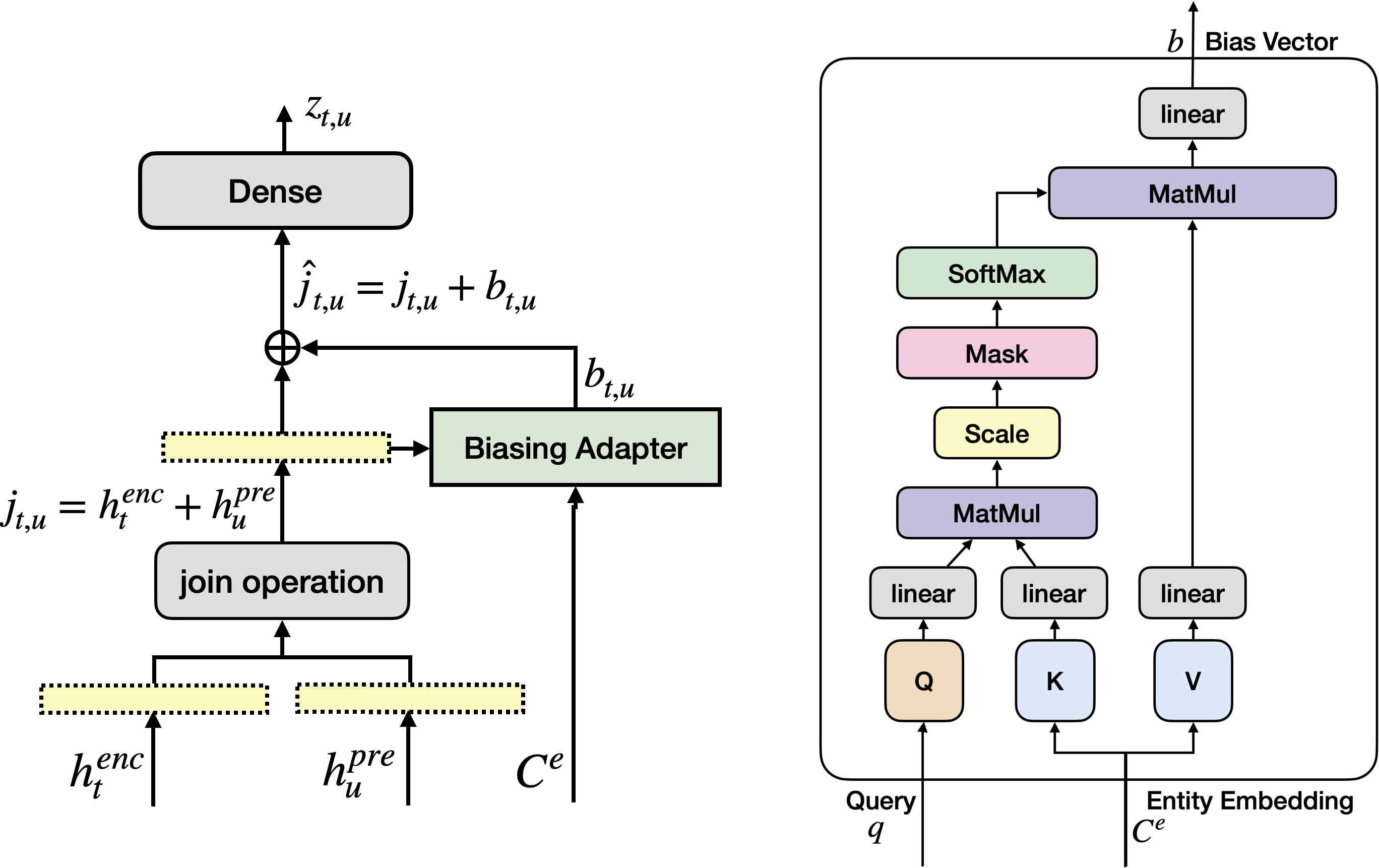}
\end{figure}
% Only captions for figure
\begin{figure}
\vspace{13pt}
  \caption{\quad (a) Joint Query \quad \quad \quad (b) Biasing Adapter} \label{fig:ad_vars}
  \vspace{-23pt}
  \end{figure}

The biasing vectors are used to update the intermediate representations of a pretrained neural transducer. Note that all updates to intermediate representations are performed via element-wise additions (denoted by $\oplus$). 
We investigate four biasing variants depending on the query being used for the biasing adapter (BA): (1) \textbf{{Enc Query:}} $\encout$ is provided as the query, and is adapted to produce $\htcap$:
\begin{align}
    \encbiasvector & = \text{BA}_{enc}(\encout, C^e); \quad \htcap = \encout \oplus \encbiasvector
    \label{eqn:1}
\end{align}
(2) {\textbf{Pred Query:}} $\decout$ is the query and adapted to produce $\hucap$:
\begin{align}
    \decbiasvector & = \text{BA}_{pre}(\decout, C^e); \quad \hucap = \decout \oplus \decbiasvector
  \label{eqn:2}
\end{align}
(3) {\textbf{Enc-Pred Query:}} Both $\encout$ and $\decout$ are adapted as in equations ~\ref{eqn:1} and ~\ref{eqn:2} before passing them to the joint network.
(4) {\textbf{Joint Query:}} The result of the join operation, $j_{t,u}$, is the query and is adapted to produce $\hat{j}_{t,u}$. This is used to update the joint representation before passing it through the activation and the dense layer. 
\begin{align*}
  \jointbiasvector & = \text{BA}_{joint}(j_{t,u}, C^e); \quad \hat{j}_{t,u} = j_{t,u} \oplus \jointbiasvector
  \end{align*}

\subsection{Adapter Style Training}
\vspace{-3pt}
By design, our \todokan{contextual} ASR model is built by auxiliary training of contextual adapters. In this style of training, all network parameters, except the adapters (shown in green in Figure~\ref{fig:RnntAdapter}), are initialized with a pretrained model trained on large amounts of generic ASR data, without any \todokan{contextextualization}. The pretrained model parameters are then kept frozen, while the adapter modules are randomly initialized and trained from scratch. Since all the updates to intermediate representations are performed by element-wise additions, the pretrained model architecture can remain the same. 
\subsection{Handling multiple catalog types}
Finally, our approach can be further extended to support multiple types of catalogs such as \todokan{Proper} Names, Device Names \todokan{(eg. `Harry Potters lamp' etc.)} \todokan{and/or Locations (eg. `bedroom', `kitchen' etc.)}. A shared catalog encoder is used to encode entities of different types. In order for the contextual adapters to distinguish between different catalog types and bias toward a specific one, we introduce a learnable `type\_embedding'. Given the entity $c_i$ and it's type $t_i \in \tau$ where $\tau$ is the set of entity types, the embedding is computed as follows. 
\begin{align*}
\cnemb = \text{Concat}(\text{BiLSTM}(\text{Embedding}(c_i)), \text{TypeEmbedding}(t_i)))
\end{align*}

%% file: sec_experiments.tex
\subsection{Datasets and Evaluation Metrics}
\label{sec:dataSet}
We use in-house voice assistant datasets with each utterance consisting of the audio and transcription randomly sampled from the VA traffic across more than 20 domains such as \textit{Communications}, \textit{Weather}, \textit{SmartHome}, and \textit{Music}. 114k hours of data are used to pre-train the baseline RNN-T and C-T models. For training the adapters, we use approximately 290 hours of data, containing a mix of \todokan{\textit{specific}} and \textit{general} training data (\textit{Mixed Dataset}) in the ratio of ($m$:1), where $m$ is a hyperparameter. \todokan{\textit{Specific}} datasets contain utterances with mentions of proper names, device names and/or locations. \textit{General} utterances are sampled from the original training data distribution. The training data is not associated with identifying information, but some utterances may contain personal information. Mixing a comparable proportion of \todokan{\textit{specific}} and \todokan{\textit{general}} utterances enables the adapters to learn when to or not to bias toward specific words. We evaluate our models and report results on two test sets - a 75 hour \textit{general} dataset, and a 20 hour \todokan{\textit{specific}} dataset \footnote{there does not exist an equivalent, publicly available \todokan{contextual} dataset}.

For our experiments, we report the relative word error rate reduction (WERR) on the \textit{general} and \todokan{\textit{specific}} sets, and the relative named entity word error rate reduction (NE-WERR) for \todokan{contextual} entity types. For WER, given a model A's WER ($\text{WER}_A$) and a baseline B's WER ($\text{WER}_B$), the WERR of A over B is computed as
$
\text{WERR} = (\text{WER}_B - \text{WER}_A)/\text{WER}_B.
$
NE-WERR is computed similarly and is used to demonstrate the improvement on \todokan{contextual} entities. Higher values indicate better performance.
\vspace{-5pt}
\subsection{Experimental Setup}
\label{sec:baseline}
We evaluate the contextual adapters with two pretrained neural transducer based ASR models, RNN-T and C-T.\\

\vspace{-3pt}
\noindent\textbf{Pretrained RNN-T and C-T.} Both RNN-T and C-T are pretrained on a large 114k hour corpus. The input audio features are 64-dimensional LFBE features extracted every 10 ms with a window size of 25 ms and resulting in 192 feature dimensions per frame. Ground truth tokens are passed through a 4000 word-piece sentence piece tokenizer \cite{sennrich-etal-2016-neural, kudo2018subword}. The RNN-T encoder network consists of 5 LSTM layers with 736 units each with a time-reduction layer (downsampling factor of 2) at layer three. The C-T encoder network consists of 2 convolutional layers with kernel size=3, strides=2, and 128 filters, followed by a dense layer to project input features to 512 dimensions. They are then fed into 12 conformer blocks~\cite{gulati2020conformer}. Each conformer block consists of a 512-node feed-forward layer, 1 transformer layer with 4 64-dim attention heads, 1 convolutional module with kernel size=32, and then feed-forward layer of 512 nodes. All the conformer blocks contain layer normalizations and residual links in between layers. All convolutions and attentions are computed on the current and previous audio frames to make it streamable. For both models, the prediction network consists of 2 LSTM layers with 736 units per layer. The outputs from the encoder and prediction network are projected through a feed-forward layer to 512 units. The joint network performs the join operation, which is a simple addition ($\oplus$). Additionally, we use a $tanh$ activation for the RNN-T model. Decoding is performed using the standard RNN-T/C-T beam search with a beam size of 8. The output vocabulary consists of 4000 word-pieces. \\

\label{subsubsec:nat}
\vspace{-3pt}
\noindent\textbf{Contextual Adapter Configuration.} The catalog encoder is a BiLSTM layer with 128 units (each for forward and backward LSTM) with an input size 64. The final output is projected to 64-dimensions. The biasing adapters project the query, key and values to 64-dimensions. The attention context vector obtained is projected to the same size as the encoder and/or prediction network or joint output sizes. For training the adapters, we use the Adam optimizer with learning rate 5e-4 trained to convergence with early stopping. We use a mix of \todokan{\textit{Specific}} and \textit{General} dataset in the ratio ($m$=1.5:1), selected based on a hyperparameter search. The contextual adapters in total make up \textless 500k parameters (\textless 1.5\% of the pretrained ASR model parameters). The maximum catalog size ($K$) is set to 300 during training to fit within memory. For experiments with multiple catalog types, we also concatenate a type embedding of size 8. We set the maximum catalog size to 300 for Proper Names, and 100 for Appliances and Device Location, respectively.\\

\vspace{-3pt}
\noindent{\textbf{Baseline -- Pretrained RNN-T/C-T.}} Our first baseline is the non-personalized pretrained RNN-T/C-T model as described previously.\\

\vspace{-3pt}
\noindent{\textbf{Basline + SF -- Pretrained RNN-T/C-T + Shallow Fusion.}} 
For the SF baseline, we build sub-word FSTs for the contextual information from user catalogs, and perform on-the-fly biasing during beam decoding, combining scores from the RNN-T model and the FST \emph{during} lattice generation \cite{rybach2017lattice,zhao2019shallow}. We perform word-piece level biasing with weight-pushing, using subtractive costs to prevent incorrectly biased sub-words that do not form full words in the FST~\cite{gourav2021personalization}.

%% file: sec_results.tex
% \usepackage{multirow}
% \usepackage{multirow}
\begin{table*}
	\caption{\small{\textbf{Results.} Relative change in WER (WERR), and \todokan{proper} name NE-WER (NE-WERR) over vanilla RNN-T and C-T models for, a) Shallow Fusion, b) Contextual Adapter (CA) variants of RNN-T and C-T models, c) Combination of Shallow Fusion with the Enc-Pred query. Note that SF weights are selected based on a hyperparameter search for each model with SF. Best results highlighted in bold.}}
	\label{tab:summary}
	\vspace{-6pt}
	\begin{tabular}{|l|c|c|c|c|c|c|}
		\hline
		\multirow{3}{*}{\textbf{Model}} & \multicolumn{3}{c|}{\textbf{RNN-T}} & \multicolumn{3}{c|}{\textbf{C-T}} \\ \cline{2-7} 
		 & \textbf{General} & \todokan{\textbf{Proper Names}} & \todokan{\textbf{Proper Names}} & \textbf{General} & \todokan{\textbf{Proper Names}} & \todokan{\textbf{Proper Names}} \\ \cline{2-7} 
		 & \multicolumn{2}{c|}{\textbf{WERR}} & \textbf{NE-WERR} & \multicolumn{2}{c|}{\textbf{WERR}} & \textbf{NE-WERR} \\ \hline
		\textbf{Baseline} & 0.00 & 0.00 & 0.00 & 0.00 & 0.00 & 0.00 \\ \hline
		\textbf{Baseline + SF} & -3.87 & +21.84 & +27.70 & -3.02 & +24.22 & +38.12 \\ \hline \hline
		\textbf{CA - Enc Query} & -1.50 & +30.37 & +33.60 & -2.01 & \textbf{+30.66} & +42.83 \\ \hline
		\textbf{CA - Pred Query} & \textbf{-1.45} & +12.43 & +13.30 & -2.42 & +11.13 & +23.34 \\ \hline
		\textbf{CA - Enc-Pred Query} & -3.12 & \textbf{+31.29} & \textbf{+34.10} & -2.87 & +30.65 & \textbf{+44.33} \\ \hline
		\textbf{CA - Joint Query} & -2.69 & +26.24 & +29.00 & \textbf{-1.14} & +30.28 & +42.83 \\ \hline \hline
		\textbf{CA - Enc-Pred Query + SF} & -4.43 & \textbf{+36.17} & \textbf{+39.70} & -2.92 & \textbf{+34.95} & \textbf{+46.68} \\ \hline
		\end{tabular}
	\vspace{-8pt}
\end{table*}
\begin{figure}[tb]
   \centering
   \includegraphics[width=0.98\linewidth]{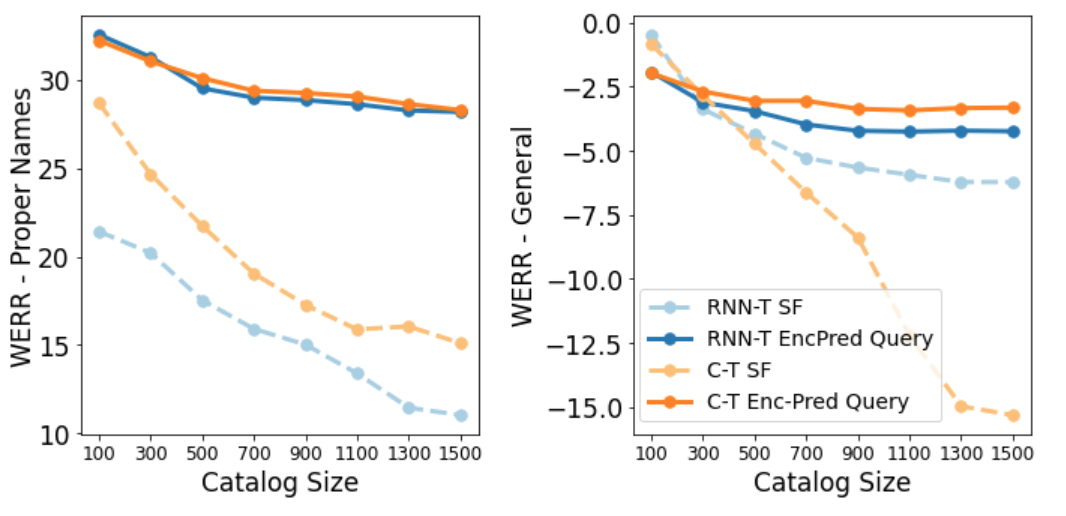}
   \caption{\textbf{WERR vs. Catalog Size.} Figure showing the WERR on \textit{General} and \todokan{\textit{Proper Names}} set for our approach (EncPred Query) vs. SF, as catalogs increase in size.}
   \label{fig:catalog_sizes}
   % \vspace{-20pt}
\end{figure}
% ---------------------- FLOAT BOX ----------------

Tables~\ref{tab:summary},~\ref{tab:training_type},~\ref{tab:ablations} present results for personalization in the \textit{Communications} domain, where \todokan{proper} names (eg. contacts) are used for biasing. Note that the baseline pretrained models work well on the \textit{General} set (below 10\% absolute WER) while poorly on the \todokan{\textit{Specific}} set (below 20\% and above 10\% absolute WER). We train contextual adapters with two types of neural transducers -- RNN-T and C-T, and demonstrate that the approach is generic and applicable to this class of models. In Table~\ref{tab:summary}, we compare contextual adapters with different query types on both -- the general test set \textit{(General)} and the \todokan{specific} \todokan{\textit{Proper Name}} test set \todokan{\textit{Proper Names}}. The proposed models outperform the baseline methods for all query types. Enc-Pred Query model achieves the greatest improvement on the \todokan{\textit{Proper Names}} set (NE-WERR of 34.1\% and 44.33\% for RNN-T and C-T, respectively). Moreover, the degradation on the \textit{General} set of our methods is smaller compared to SF (-1.14\% vs -3.02\% WERR for C-T). Interestingly, we found that adapting the encoder outputs (as in Enc and Enc-Pred Query) is more important than adapting only the prediction network outputs (Pred Query). Finally, we show that combining contextual adapters with SF, we obtain the most improvement on \todokan{\textit{Proper Names}} test set (39.70\% and 46.68\% NE-WERR for RNN-T and C-T), showing that these two approaches are complementary to each other. We do notice a slighly higher degradation on the General set. We hypothesize that this is due to overbiasing from using both approaches.\\
\noindent\textbf{Importance of Adapter-Style Training.}
% Training -- AdT vs FFT
We observe that it is essential to freeze the base RNN-T parameters to obtain good improvements on \todokan{proper} names (Table~\ref{tab:training_type}). 
Fine-tuning all parameters on the 290 hour \textit{Mixed dataset} degrades the performance on the \textit{General} set significantly (-19.23\% WERR) resulting from catastrophic forgetting\cite{kirkpatrick2017overcoming} as the base model parameters get updated. We hypothesize that this is due to the distribution shift in the \textit{Mixed Dataset} causing the model to lose performance on the dataset it was original trained on (i.e the General dataset). Adapter style training can effectively reduce catastrophic forgetting (-19.23 vs -3.11 WERR). Furthermore, it outperforms full fine-tuning on the \todokan{\textit{Proper Names}} set by a large margin +31.29\% vs. +2.19\% WERR.

\begin{table}[t]
	\vspace{-6pt}
	\resizebox{\textwidth}{!}{%
	\begin{tabular}{|l|c|c|}
		\hline
		\textbf{Training Type} & \textbf{General} & \todokan{\textbf{Proper Names}} \\ \hline
		Adapter Training (Base RNN-T Frozen) & -3.11 & +31.29 \\ \hline
		Full fine-tuning (Base RNN-T Updated) & -19.23 & +2.19 \\ \hline
		\end{tabular}}%
	\caption{\small{\textbf{Adapter-style training vs. Full fine-tuning.} Impact of freezing vs. unfreezing the base RNN-T parameters on WERR for \textit{General} and \todokan{\textit{Proper Names}} Test set compared to baseline model.}}
	\label{tab:training_type}
\end{table}
\noindent\textbf{Importance of the \nobias token and biasing adapters.}
With the best-performing contextual adapter model using Enc-Pred Query in Table~\ref{tab:summary}, we see that removing the \nobias token results in a significant WER degradation (Table~\ref{tab:ablations}) on the General set (-15.31\% vs -3.11\% WERR). It indicates the \nobias token is essential to help the adapters learn when not to bias. Further, to verify if the biasing adapters really learn to attend towards the correct \todokan{proper} names, we train a contextual adapter with random word pieces in the catalog. We see a huge drop in WERR on the \todokan{\textit{Proper Names}} set (31.29\% to 6.82\%) indicating that the biasing adapters have learned where to bias toward when the right \todokan{proper} names are provided.
\begin{table}[t]
	\vspace{-6pt}
	\begin{tabular}{|l|c|c|}
		\hline
		\textbf{Ablations} & \textbf{General} & \todokan{\textbf{Proper Names}} \\ \hline
		Enc-Pred Query & -3.11 & +31.29 \\ \hline
		\quad \quad  without \nobias & -15.31 & +31.29 \\ \hline
		\quad \quad  with Random Catalog & -2.87 & +6.82 \\ \hline
		\end{tabular}
		
	\caption{\small{\textbf{Ablations.} Effect of removing \nobias and effect of providing random input to catalog encoder.}}
	
	\label{tab:ablations}
\end{table}

\begin{table}[t]
	\vspace{-6pt}
	\resizebox{0.55\textwidth}{!}{%
	\begin{tabular}{|l|c|l|}
\hline
\textbf{Entity type} & \textbf{w/ TE} & \textbf{w/o TE} \\ \hline
Appliance & +25.00 & +25.00 \\ \hline
Location & +29.17 & +29.17 \\ \hline
Proper Names & +38.66 & +27.53 \\ \hline
\end{tabular}}%
	\caption{\small{\textbf{Multiple Catalogs.} NE-WERR for different types with and without type embedding (TE).}}
	\label{tab:multiple_types}
	\end{table}

\noindent\textbf{Impact of catalog sizes.}
We further evaluate the robustness of contextual adapters towards large catalogs where most entities are irrelevant and do not appear in the reference transcript. In Fig.~\ref{fig:catalog_sizes}, we show WERRs for the Enc-Pred Query variant over baseline RNN-T/C-T. In \todokan{\textit{Proper Names}} set, our approach is more robust than SF scales better; the WERRs remain above 25\% even at large sizes. The gap between SF and our approach increases as the catalog sizes increase. Additionally, our method degrades lesser on the \textit{General} set when compared to SF as catalogs grow in size.

\noindent\textbf{Handling multiple catalog types.}
Finally, we show that our method can handle multiple catalog types by incorporating the catalog type embedding (TE), and evaluated on $\sim$20 hours of \todokan{contextual} utterances for each entity type -- \textit{Appliance}, \textit{Location} and \todokan{\textit{Proper Names}}. As seen in Table~\ref{tab:multiple_types}, our method learns which type of catalog to bias and achieves consistent improvements not only for \todokan{proper} names, but also for Appliance and Location. Further, removing the TE degrades the performance on proper names showing it's importance. 